**Title:**

Development of an Extractive Clinical Question Answering Dataset with Multi-Answer and Multi-Focus Questions

**Authors:**

[1]Sungrim Moon Ph.D.

[1]Huan He Ph.D.

[1,2]Hongfang Liu Ph.D.

[1,2,3]Jungwei W. Fan Ph.D.*

[1]Department of Artificial Intelligence & Informatics

[2]Center for Clinical and Translational Science

[3]Department of Quantitative Health Sciences

Mayo Clinic, Rochester, United States

*Corresponding author: fan.jung-wei@mayo.edu

200 1st Street SW, RO_HA_07_741B-I, Rochester, MN 55905

(507) 538-1191

**Word count:** 3,731 words (from Introduction to Conclusion)

## Abstract

**Background**

Extractive question-answering (EQA) is a useful natural language processing (NLP) application for answering patient-specific questions by locating answers in their clinical notes. Realistic clinical EQA can have multiple answers to a single question and multiple focus points in one question, which are lacking in the existing datasets for development of artificial intelligence solutions.

**Objective**

Create a dataset for developing and evaluating clinical EQA systems that can handle natural multi-answer and multi-focus questions.

**Methods**

We leveraged the annotated relations from the 2018 National NLP Clinical Challenges (n2c2) corpus to generate an EQA dataset. Specifically, the 1-to-N, M-to-1, and M-to-N drug-reason relations were included to form the multi-answer and multi-focus QA entries, which represent more complex and natural challenges in addition to the basic one-drug-one-reason cases. A baseline solution was developed and tested on the dataset.

**Results**

The derived RxWhyQA dataset contains 96,939 QA entries. Among the answerable questions, 25% require multiple answers, and 2% ask about multiple drugs within one question. There are frequent cues observed around the answers in the text, and 90% of the drug and reason terms occur within the same or an adjacent sentence. The baseline EQA solution achieved a best f1-measure of 0.72 on the entire dataset, and on specific subsets, it was: 0.93 on the unanswerable questions, 0.48 on single-drug questions versus 0.60 on multi-drug questions, 0.54 on the single-answer questions versus 0.43 on multi-answer questions.

Discussion

The RxWhyQA dataset can be used to train and evaluate systems that need to handle multi-answer and multi-focus questions. Specifically, multi-answer EQA appears to be challenging and therefore warrants more investment in research.

Conclusions

We created and shared a clinical EQA dataset with multi-answer and multi-focus questions that would channel future research efforts toward more realistic scenarios.

# Introduction

## Background

The thought process involved in clinical reasoning and decision-making can be naturally framed into a series of questions and answers [1, 2]. Achieving human-like question-answering (QA) capability is highly regarded in artificial intelligence (AI). Medical QA research has garnered terrific momentum over the past decade, and a new generation of AI scientists is updating state-of-the-art at a daunting pace almost every month, if not week. One of the very sought-after applications is to find the answer within a given document, or extractive question-answering (EQA), which enables patient-specific QA based on the information mentioned in the clinical text [3]. As an essential component in most AI engineering, EQA training data not only determines the likelihood of success in terms of annotation quality but also the fidelity of representing the target scenario.

Along with other issues observed in existing medical EQA corpora [4], the mainstream annotation approach knowingly simplifies the task into a "one answer per document" scheme. Although the simplification makes development and evaluation easier for promoting the initial growth of the field, it is unrealistic because EQA can naturally have multiple qualified answers (or answer components) within one document, and often all of them must be captured to sufficiently answer a question [5]. Moreover, a question can naturally involve multiple focus points like "Why A, B, and C…" rather than requiring the user to ask one question for each point. To address this gap, we created an EQA dataset that involves realistic, multi-answer and multi-focus cases by converting the concept-relation annotations from an existing clinical natural language processing (NLP) challenge dataset. Our generated RxWhyQA dataset includes a total of 96,939 QA entries, where 25% of the answerable questions require the identification of multiple answers and 2% ask about multiple drugs within one question. We also developed a baseline solution for multi-answer QA and tested it on the RxWhyQA.

The novelty of the study was reframing the original relation identification task into an EQA task, which simplifies the conventional two-step approach of named entity recognition plus relation classification into one-step information extraction guided by natural language questions. Our primary contribution is the RxWhyQA as a resource that offers realistic constructs to facilitate NLP research in this underexplored area. To our knowledge, there has not been any EQA dataset that contains multi-answer and multi-focus questions based on clinical notes.

Related work

QA is a versatile task that can subsume diverse NLP tasks when properly represented [6]. The extractive question-answering (EQA) [7] task in NLP has more than a decade of research. As the name implies, EQA can be viewed as question-guided information extraction from a given text. Unlike conventional approaches that require the identification of the involved entities as one task followed by determining the target relation between the entities as the other task, EQA consolidates these steps into a smooth one-shot task where the user asks a natural language question for the system to understand the focus point, relevant cues in the text, and locate the answer that satisfies the relation of interest. Although EQA demands higher machine intelligence, it is efficient in terms of the data schema for modeling and the human-computer interaction for users.

The Stanford Question Answering Dataset (SQuAD) [8] established a widely adopted framework for EQA, and in the later version 2.0 [9], the task also requires a system to refrain from answering when no suitable answer is present in the text. In the clinical domain, corpora have been developed for EQA based on electronic health records (EHR). In Raghavan et al. [10], medical students were presented with structured and unstructured EHR information about each patient to come up with realistic questions for a hypothetical office encounter. Using the BioASQ dataset based on biomedical literature, Yoon et al. [5] proposed a sequence tagging approach to handling multi-answer EQA. In the consumer health domain, Zhu et al. [11] developed a Multiple Answer Spans Healthcare Question Answering (MASH-QA) dataset specifically involving multiple answers of non-consecutive spans in the target text. As a non-English example, Ju et al. [12] developed a Conditional Multiple-span Chinese Question Answering dataset from an online QA forum. Pampari et al. [13] developed the emrQA, a large clinical EQA corpus generated through template-based semantic extraction from the i2b2 NLP challenge datasets. We took a similar approach as emrQA, but additionally included multi-answer and multi-focus questions that better reflect natural clinical EQA scenarios.

## Methods

### Generating the QA annotations from a relation identification challenge

Our source data was based on the annotations originally created for the National NLP Clinical Challenges (n2c2) of 2018, which aimed to identify adverse drug events (ADEs) by extracting various drug-related concepts and classifying their relations in the clinical text [14]. Their final gold standard included 83,869 concepts and 59,810 relations in 505 discharge summaries. In this study, we focused on generating QA pairs from the subset of drug and reason concepts (i.e., mainly about the prescribing justification) and the relations between the concepts. Each relation consisted of two arguments: a drug concept and a reason concept, as in an example pair like *drug-reason* (morphine, pain). Accordingly, we could derive a question around the drug concept: "Why was morphine prescribed to the patient?" and the reason concept "pain" would be designated as the answer. In the n2c2 corpus, each pair of drug and reason concepts had their text mentions annotated in the corresponding clinical document. The properties make a good EQA dataset where the system is expected to consider the actual contexts surrounding the drug and reason rather than performing a simple lookup. This is especially important for extracting off-label uses because a standard indication knowledge base would not cover those exceptions documented in the real-world clinical text.

From the n2c2 annotations on each clinical document, we leveraged several relation types between the drug and reason concepts: 1 drug 0 reason, 1 drug 1 reason, 1 drug N reasons, N drugs 1 reason, or M drugs N reasons. The most straightforward was the 1 drug 1 reason relations (e.g., the morphine-pain relation above), each translated into a 1-to-1 QA entry. The 1 drug 0 reason relations apparently corresponded to the 1-to-0 (unanswerable) QA entries. We preserved the 1 drug N reasons directly as 1-to-N QAs that require locating multiple answers in the text. For the N drugs 1 reason and M drugs N reasons, we preserved the original multi-drug challenge in the questions, as in "Why were amlodipine, metoprolol, and isosorbide prescribed to the patient?" The M drugs N reasons would also derive multi-answer entries like those derived from 1 drug N reasons. On top of the generated QA entries, we also supplemented paraphrastic questions [15] that may enhance the generalizability of the trained systems.

## Quantitative and qualitative analysis of the derived QA annotations

Along with descriptive statistics of the QA entries and the number of answers per question, we computed frequencies of the specific drug and reason concept terms (after applying lexical normalization such as lowercase) among the QA entries. The frequencies were meant to offer an intuitive estimate of the abundance of train/test data available for each specific concept or concept pair. We then randomly sampled 100 QA entries for manual review: 50 from those with a single answer and 50 from those with multiple answers. The common patterns informative to QA inference were summarized, offering a clue on what the potential AI solutions could leverage. In addition, we measured the distance (by the number of sentences) between the question and answer concepts. For each specific drug-reason pair, we took the shortest distance if there were multiple occurrences of either concept. The distance was deemed 0 if the pair occurred within the same sentence. The distance may serve as a surrogate for measuring the challenge to AI systems, where a longer distance implies a more challenging task.

## Development of a baseline solution

### Data preparation and model training

The annotations conform to the SQuAD 2.0 JSON format and can be readily used to train Bidirectional Encoder Representations from Transformers (BERT) [16] for EQA tasks. We randomly partitioned the dataset into the train, develop (dev), and test sets by the 5:2:3 ratio, corresponding to 153, 50, and 100 clinical documents, respectively. The random split was done three times, each executed as a separate run of the experiment for quantifying performance variability. The base language model was ClinicalBERT [17], a domain-customized BERT trained on approximately two million clinical documents from the MIMIC-III v1.4 database. We fine-tuned ClinicalBERT first on a why-question subset of the SQuAD 2.0, followed by fine-tuning on the train set. Training parameters used in the ClinicalBERT fine-tuning were batch_train_size=32, max_seq_length=128, doc_stride=64, learning_rate=3e-5, and epochs=5. The dev set was then used to learn the threshold for determining when the ClinicalBERT model should refrain from giving any answer.

*Incremental masking to generate multiple answers*

To force the fine-tuned ClinicalBERT model to continue seeking other suitable answers in each clinical document, we implemented the following process on the test set as a heuristic baseline:

1. Let the EQA model complete its usual single-answer extraction and record the string of the top answer. No further action was needed if the model refrained from answering.
2. Performed case-insensitive string search using the top answer throughout the clinical note, where it was extracted and replaced every occurrence into a dummy underscore "______" string of identical length. This literally generated a new version of the text by masking the original top answer in each question.
3. Ran the same EQA model for another round on the entire masked test set again to see whether the model was able to identify additional answers elsewhere or started to refrain from answering.

The three steps above were repeated until the model did not generate any new answers on the entire test set. Together the model training and the heuristic multi-answer generation process are summarized in **Figure 1**.

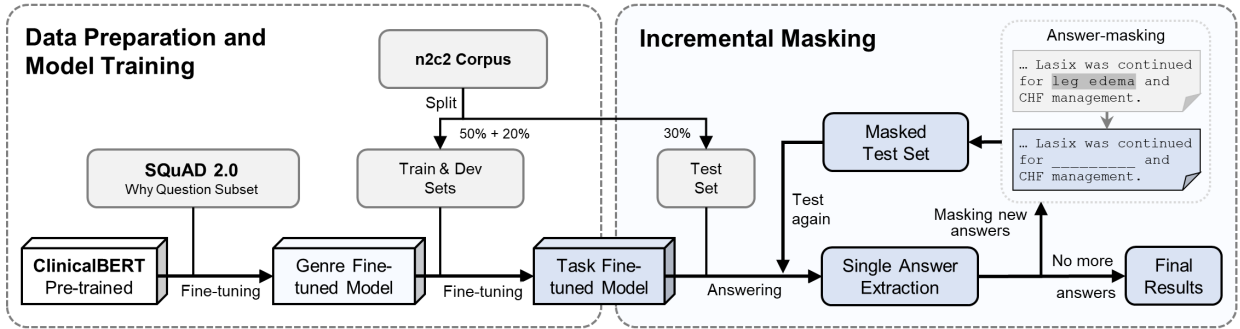

**Figure 1.** A flowchart of our heuristic approach to making a single-answer EQA model generates multiple answers by incremental masking. The main steps go from left to right. The upper right "Answer-masking" box illustrates an example of the masking where the model's answer "leg edema" is replaced with a dummy underscore to force the model to look for viable alternative answers elsewhere in the text.

### Evaluation of the baseline solution

After the first round of masking, we began to have more than one answer generated by the model for some of the questions. Accordingly, the evaluation program (specifically for the overlap mode) was adapted to accommodate such M-to-N answer comparisons in determining the token-wise proportional match. When anchoring on each gold answer, we picked the model answer with the most overlapping tokens as the best answer in setting the weighted true positive (TP) and false negative (FN); the weighted false positive (FP) was set vice versa by anchoring on each model answer; see **Eq.(1)** to **Eq.(4)** for definitions. On top of these weighted matches between gold and model answers in each question, we tallied them over the entire test set to compute the solution's precision, recall, and f1-measure, followed by qualitative error analysis.

$$\text{weighted } TP_{gold} = \frac{\# \text{ of overlap tokens with the best model answer}}{\# \text{ of tokens in the gold answer}} \qquad \text{Eq.(1)}$$

$$weighted\ FN = 1 - weighted\ TP_{gold} \qquad Eq.(2)$$

$$weighted\ TP_{model} = \frac{\#\ of\ overlap\ tokens\ with\ the\ best\ match\ gold\ answer}{\#\ of\ tokens\ in\ the\ model\ answer} \qquad Eq.(3)$$

$$weighted\ FP = 1 - weighted\ TP_{model} \qquad Eq.(4)$$

## Results

### Descriptive statistics of the derived RxWhyQA dataset

We leveraged a total of 10,489 relations from the n2c2 ADEs NLP challenge and derived the dataset, consisting of 96,939 QA entries. **Table 1** summarizes the five major drug-reason relation categories in the n2c2 corpus, the strategies that we implemented to convert them into QA entries, and their resulting frequencies. **Figure 2** shows the distribution for the number of answers per question: 75% of the questions have a single answer, while 25% of them require multiple answers. Note that duplicate answer terms located at different positions of the clinical documents were retained. For example, the procedure "CT" might occur at several places in the text and be recorded as the answer to "Why was the patient prescribed contrast?" We included each such identical term and their different offset as multiple answers so that the EQA solutions may leverage such nuances. The final dataset was formatted into a SQuAD-compatible JSON file and shared through the n2c2 community annotations repository.[a] **Figure 3** illustrates a multi-answer entry in the RxWhyQA dataset.

---

[a] https://portal.dbmi.hms.harvard.edu/projects/n2c2-du/

**Table 1.** Categories, examples, and conversion strategies for making the drug-reason relations into the EQA annotations.

| Category in the n2c2 corpus | Example | Conversion strategy | Number of entries |
|---|---|---|---|
| 1 *Drug*, no **Reason** | *Mirtazapine* 15 mg PO QHS. [**only the drug is mentioned but no reason is documented**] | Make an unanswerable QA entry | 46,278 |
| 1 *Drug*, 1 **Reason** | The patient received *morphine* for **pain** as needed. | Make a 1-to-1 QA entry | 28,224 |
| N *Drugs*, 1 **Reason** | **Hypertension**: Severely elevated blood pressure. Started *amlodipine*, *metoprolol*, and *isosorbide*. | Break into N separate 1-to-1 relation and make each a 1-to-1 QA entry | |
| 1 *Drug*, N **Reasons** | *albuterol sulfate* 90 mcg… Puff Inhalation Q4H for **sob** or **wheeze**. | List the N reasons under answer block to form a 1-to-N QA entry | 22,437 |
| M *Drug*, N **Reasons** | **Left frontoparietal stroke** - maintained on *ASA* and *plavix* …. Hx of **CVA**: re-started *ASA/Plavix* as per GI team's recs. | List the N reasons under answer block to form an M-to-N QA entry | |

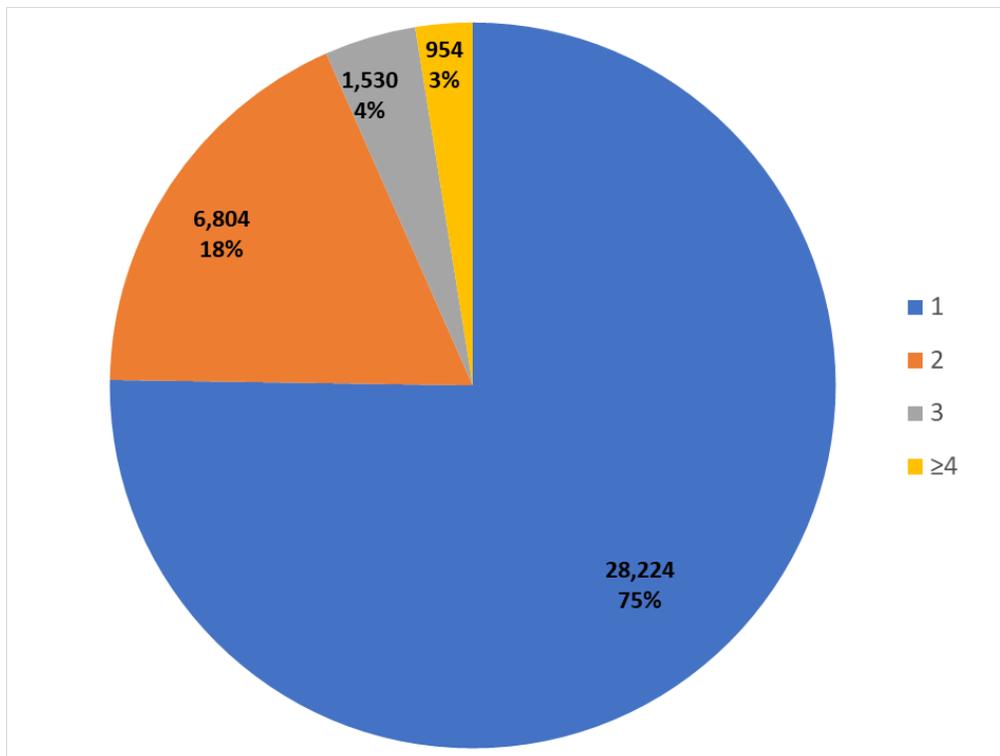

**Figure 2.** Frequency and percentage for the number of unique answers among those answerable questions. For example, there are 954 questions that have 4 or more answers (dark yellow).

```
{
    "question_template": "Why was the patient prescribed |medication|?",
    "question": "Why was the patient prescribed Metoprolol?",
    "id": "141586.xml_M100_3",
    "_mname": "Metoprolol",
    "answers": [
        {
            "text": "Atrial fibrillation",
            "answer_start": 8695
        },
        {
            "text": "Hypertension",
            "answer_start": 9115
        }
    ],
    "is_impossible": false
}
```

**Figure 3.** A multi-answer entry in the generated RxWhyQA dataset. The "id" field is the unique ID for the QA entry in the dataset. The "_mname" field indicates the medication name, i.e., the anchor concept in the question. The "answer_start" is the character offset where the answer term occurs in the clinical document, which is hosted in the "context" field (not shown here). When "is_impossible" is false, which means this is an answerable QA entry.

Content analysis of the RxWhyQA dataset

The five most frequently asked drug terms (with noting the number of QA entries) in the answerable questions (frequencies) are: coumadin (1,278), vancomycin (1,170), lasix (963), acetaminophen (801), and antibiotics (783). Without any overlap, the five most frequent drug terms in the unanswerable questions are: docusate sodium (648), metoprolol tartrate (504), aspirin (468), pantoprazole (450), and penicillins (414). Among the answerable QA entries, the five most frequently seen pairs are: acetaminophen-pain (504), senna-constipation (369), oxycodone-pain (261), coumadin-afib (252), and acetaminophen-fever (234). As a potential surrogate measure of the task difficulty, **Figure 4** shows the distribution for the number of sentences between the question anchor and answer term in each answerable QA entry. The majority (72%, n=32,409) of the drug and reason terms occur within the same sentence, and the portion increases to 90% (72%+18%) when adding those with the drug and reason occurring in an adjacent sentence (i.e., distance=1). In the extreme case, the drug and reason terms are 16 sentences apart from each other. **Table 2** summarizes the commonly observed contexts from manually reviewing 100 random samples of the answerable QA entries.

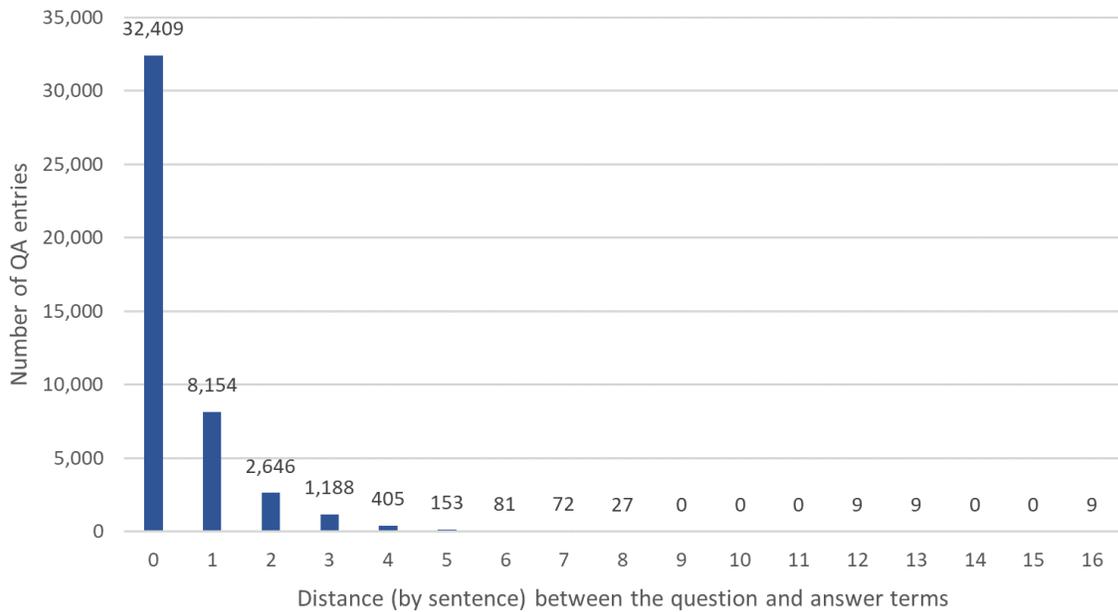

**Figure 4.** Distribution for the distance between the question and answer terms. A distance of 0 indicates that the question and answer terms occur within the same sentence.

**Table 2.** Common patterns (observed >10 times) between the question and the answer terms in 100 random QA entries. Each Reason or Drug represents where a question or answer anchor term occurs in the pattern. The shorthands are used as follows: ellipsis stands for zero to multiple words, round brackets are for scoping, square brackets with pipes indicate a boolean OR set, and a question mark is a binary quantifier for pr
esence or absence.

| Pattern | Frequency |
| --- | --- |
| **Reason** … (being)? [received\|started\|restarted\|required\|maintained\|continued?] (on)? *Drug* | 25 |
| *Drug* … [prn\|PRN\|(as needed for)?] **Reason** | 18 |
| *Drug* … (was)? [attempted\|given\|dosing\|taking] for (any)? [possible\|likely\|presumed]? **Reason** | 14 |
| **Reason** … (was)? [managed\|treated\|improved\|recommended\|downtrended\|resolved\|reversed\|needed] with *Drug* | 13 |

### F1-measure of the baseline EQA solution

The performance in the f1-measure across three experiment runs is summarized in **Figure 5**, where the sub-figures represent different slices. Specifically, the underlying set relations are: The full set **(a)** minus the unanswerable questions **(b)** gives the answerable questions, which can be represented by either **(c)** plus **(d)** if sliced per the number of answers or by **(e)** plus **(f)** if sliced per the number of drugs asked in the question. Each bar represents the average f1 value across the runs and with the range marked for each incremental masking step. We can see in **(a)** that the overall f1 ascended right after applying the first round of answer masking (from "original" to "mask 1", $p<0.05$), which then stays flat throughout the remaining mask iterations. The increase of f1 in **(a)** corresponds to the exact pattern in **(d)**, suggesting that the performance gain was mainly from the multi-answer questions, i.e., the target originally intended by the masking. Multi-answer questions appear to be more challenging than single-answer questions comparing **(c)** and **(d)**. According to **(e)** and **(f)**, asking about multiple drugs at once made it easier for the model to find the right answer, but also with wide performance variance. The BERT model was good at refraining from answering unanswerable questions, as indicated by the high f1 values in **(b)**. The detailed results of the three experiment runs are available in the **Supplement File.**

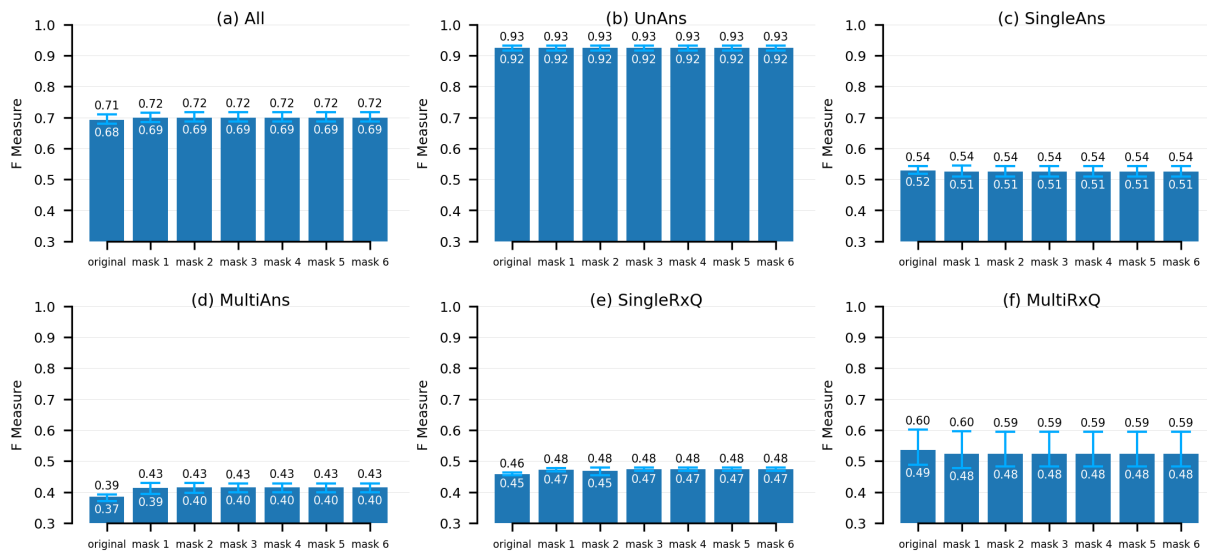

**Figure 5.** F1-measures of the fine-tuned BERT EQA model across the incremental masking rounds. Each bar represents the average f1-measure based on three experiment runs, with the min and max range marked (light blue). Subset legends: (a) the full set, (b) unanswerable questions, (c) questions with exactly one answer, (d) questions with multiple answers, (e) questions asking about a single drug, (f) questions asking about multiple drugs.

## Discussion

### Significance and contributions

Although why-question answering only covers a subdomain of clinical QA, it represents a unique category that deals with the cause, motivation, circumstance, and justification. It was estimated that 20% of the top ten question types asked by family physicians [18] could be rephrased into a why-question. Clinical whyQA is important because: 1) the ultimate task resembles expert-level explanatory synthesis of knowledge and evidence, and 2) it aligns with identifying reasons for the decisions documented in clinical text. Therefore, the contents and challenges offered by the RxWhyQA dataset itself have independent, practical value for developing clinical QA applications.

The generated RxWhyQA dataset can serve as the training and testing of AI systems that target excerpting pertinent information in a clinical document to answer patient-specific questions. In addition to the unanswerable questions that require a system to refrain from extracting false positive answers, the RxWhyQA features 9,288 questions that require the system to identify multiple answers, a realistic challenge in clinical QA. The dataset also contains 611 questions that ask about the reason for prescribing multiple drugs at once. The multi-answer and multi-focus questions represent a key improvement beyond existing clinical EQA datasets, of which the rigid constructs would preclude AI solutions from learning to deal with more realistic use scenarios. Additionally, our experiments on these special constructs validated the challenging nature of multi-answer questions and revealed that multi-focus questions may turn out to be easier due to richer information for use by the model.

## Properties found about the RxWhyQA dataset

The frequent drugs and drug-reason pairs likely imply the clinical practice in the original n2c2 cohort. The finding that the top five drugs in the unanswerable questions (i.e., no answer provided in the gold annotation) were different from that in the answerable questions suggests that the prescription of certain drugs might be self-evident without needing a documented reason. Our question-answer mentioning distance analysis showed that 90% of the drug-reason pairs were within the same or an adjacent sentence in the RxWhyQA dataset, indicating modest demand for long-distance inference by AI solutions. We were able to identify frequent contextual patterns such as "PRN" (pro re nata) or "as needed for" (**Table 2**) that AI models may learn to facilitate locating the answers.

## Behavior of the baseline EQA solution

The notable increase of f1-measure in **Figure 5(d)** after applying one round of masking suggests that the masking effectively forced the BERT model to look elsewhere that hit additional answers, despite the benefit soon plateaus thereafter (might be immediately hitting the majority of the second answer, per **Figure 2**). Interestingly, we noticed in many cases that the model clung on to the masked span (i.e., capturing the "________" as an answer) where some of such strong contextual patterns were present. This phenomenon supports that transformers-based EQA models do leverage context information than merely memorizing the surface question-answer pairs. Moreover, our post hoc inspection noted correct (synonymous) answers found by the model that were not in the gold annotation (e.g., "allergic reaction" versus "anaphylaxis" to a question about "epipen"), suggesting that the performance could be underestimated. Of caveat, we were aware that our baseline solution was essentially a convenient hack that made a model trained for single-answer EQA to find multiple answers through a stepwise probing procedure. As more advanced approaches constantly emerge [19, 20], we welcome the research community to evaluate them by using the RxWhyQA dataset.

## Limitations

We admit several limitations in the study: 1) the source n2c2 corpus represented a specific cohort that may not generalize to every clinical dataset, 2) we did not exhaustively diversify the paraphrastic questions but left it to future exploration of other promising approaches [21], 3) we did not intend to extensively compare state-of-the-art solutions for multi-answer QA but to offer a convenience baseline along with releasing the RxWhyQA corpus, 4) the drug-reason relations represent a narrow topic for EQA development and evaluation. However, we believe that the definite theme would preferably make it a less confounded test set for assessing the effect of multi-answer and multi-focus questions on AI systems.

## Conclusions

We derived and shared the RxWhyQA, an extractive question-answering dataset for training and testing systems to answer patient-specific questions based on clinical documents. The RxWhyQA includes 9,288 multi-answer questions and 611 multi-focus questions, each representing a critical scenario not well covered by existing datasets. Upon evaluating a baseline solution, the multi-answer questions did appear to be more challenging than single-answer questions. Although the RxWhyQA focuses on why-questions derived from drug-reason relations, it offers a rich dataset involving realistic constructs and exemplifies an innovation in recasting NLP annotations of different tasks for EQA research.

## Funding Statement

The research was supported by the National Center for Advancing Translational Sciences U01TR002062.

## Acknowledgements

We thank the n2c2 organizers for making the annotations available to the research community. The study was partly supported by the Mayo Clinic Kern Center for the Science of Health Care Delivery.

## Conflicts of Interest

None declared.

## Authors' Contributions

JWF conceived the study. HL offered scientific consultation. SM implemented the data conversion and analysis. HH assisted in the data conversion and graphic presentation. JWF and SM drafted the manuscript. All authors contributed to the interpretation of the results, critical revision of the manuscript, and approved the final submission.

## Abbreviations

ADE: adverse drug event

AI: artificial intelligence

BERT: Bidirectional Encoder Representations from Transformers

EHR: electronic health records

EQA: extractive question-answering

n2c2: National NLP Clinical Challenges

NLP: natural language processing

SQuAD: Stanford Question Answering Dataset